\documentclass[journal]{IEEEtran}
\usepackage{amsmath,amsfonts}
\usepackage{algorithmic}
\usepackage{array}
\usepackage[caption=false,font=normalsize,labelfont=sf,textfont=sf]{subfig}
\usepackage{textcomp}
\usepackage{stfloats}
\usepackage[numbers]{natbib} % 或用 \usepackage{cite}
\usepackage{url}
\usepackage{verbatim}
\usepackage{graphicx}
\usepackage{booktabs}
\usepackage{multirow}
\usepackage{balance}
\usepackage{hyperref} 
\def\BibTeX{{\rm B\kern-.05em{\sc i\kern-.025em b}\kern-.08em
    T\kern-.1667em\lower.7ex\hbox{E}\kern-.125emX}}
\usepackage{balance}
\begin{document}
\title{DiffusionUavLoc: \,Visually Prompted Diffusion for Cross-View UAV Localization}
\author{Tao Liu, Kan Ren and Qian Chen
\thanks{This work was supported in part by the National Natural Science Foundation of China under Grant 62175111, and the Fundamental Research Funds for the Central Universities under Grant No. 30922010715. (Corresponding author: Kan Ren).}
\thanks{Tao Liu, Kan Ren and Qian Chen are with School of Electronic and Optical Engineering, Nanjing University of Science and Technology, Nanjing, 210094, Jiangsu, China. 
(e-mail: liutao23@njust.edu.cn; k.ren@njust.edu.cn; chenq@njust.edu.cn)}}

\markboth{Journal of \LaTeX\ Class Files,~Vol.~18, No.~9, September~2020}%
{How to Use the IEEEtran \LaTeX \ Templates}

\maketitle

\begin{abstract}
With the rapid growth of the low-altitude economy, unmanned aerial vehicles (UAVs) have become key platforms for measurement and tracking in intelligent patrol systems. However, in GNSS-denied environments, localization schemes that rely solely on satellite signals are prone to failure. Cross-view image retrieval-based localization is a promising alternative, yet substantial geometric and appearance domain gaps exist between oblique UAV views and nadir satellite orthophotos. Moreover, conventional approaches often depend on complex network architectures, text prompts, or large amounts of annotation, which hinders generalization. To address these issues, we propose DiffusionUavLoc, a cross-view localization framework that is image-prompted, text-free, diffusion-centric, and employs a VAE for unified representation. We first use training-free geometric rendering to synthesize pseudo-satellite images from UAV imagery as structural prompts. We then design a text-free conditional diffusion model that fuses multimodal structural cues to learn features robust to viewpoint changes. At inference, descriptors are computed at a fixed time step t and compared using cosine similarity. On University-1652 and SUES-200, the method performs competitively for cross-view localization, especially for satellite-to-drone in University-1652.Our data and code will be published at the following URL: https://github.com/liutao23/DiffusionUavLoc.git.

\end{abstract}


\section{Introduction}
\IEEEPARstart{W}{ith} the growth of the low-altitude economy, unmanned aerial vehicles (UAVs) have become indispensable tools for measurement and tracking in intelligent patrol systems~\cite{r1}. Most current UAVs rely on the Global Navigation Satellite System (GNSS) for positional signals; however, GNSS is fragile and susceptible to jamming and spoofing. Conventional solutions typically combine an inertial navigation system (INS) with additional sensors to achieve relative localization, but such methods suffer from error accumulation that leads to drift~\cite{r2}. In recent years, fueled by advances in deep learning and the emergence of large UAV localization datasets~\cite{r3,r4,r5,r6,r7}, visual geolocalization has become a powerful alternative~\cite{r8}. This paradigm formulates UAV localization as an image retrieval and matching problem: the UAV matches real-time forward-looking images captured by its onboard camera (query images) against geo-referenced satellite tiles (reference images) to determine its location.

\begin{figure*}
  \centering
  \includegraphics[width=\textwidth]{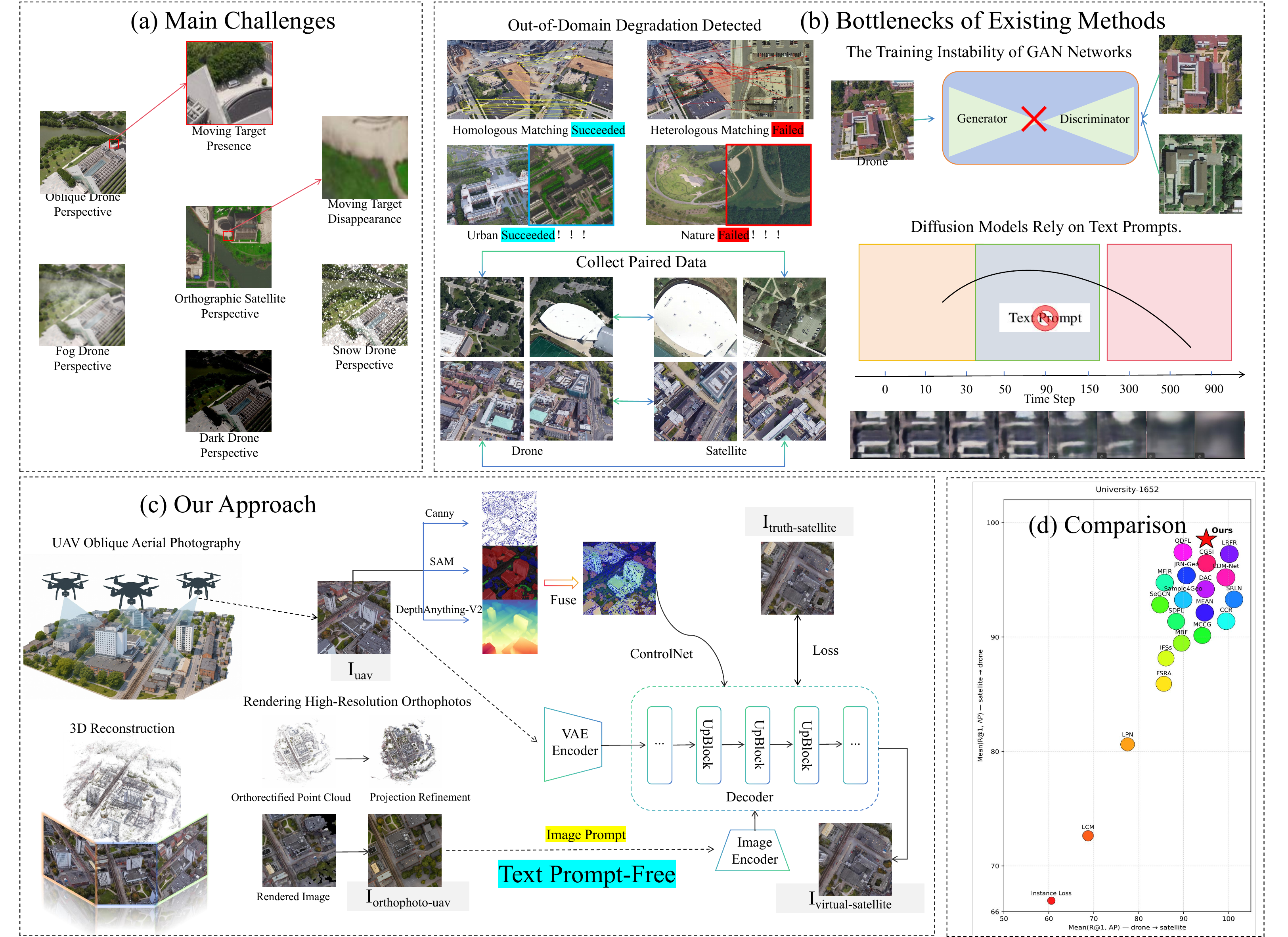}
  \caption{\textbf{Motivation and overview of \emph{DiffusionUavLoc}.}
  \textbf{(a) Main challenges.} Cross-view UAV–satellite matching suffers from large gaps in imaging geometry and appearance: oblique, low-altitude UAV views versus near-nadir satellite orthophotos, scale and occlusion changes, adverse conditions (fog, darkness, snow), and the presence or absence of moving targets. 
  \textbf{(b) Bottlenecks of existing methods.} These factors cause out-of-domain degradation, where homologous urban pairs are matched while heterogeneous natural scenes often fail; representative labeled drone–satellite pairs are illustrated. GAN-based translation is training-unstable, and standard diffusion pipelines rely on text prompts, which provide weak geometric guidance and yield blurry or misaligned cross-view correspondences across timesteps.
  \textbf{(c) Our approach.} We render high-resolution orthophotos from oblique UAV imagery via training-free geometric orthorectification and projection refinement to produce pseudo-satellite image prompts. Multimodal structural cues (e.g., Canny edges, SAM masks, and DepthAnything-v2 depth) are fused and injected through ControlNet into a text-free conditional diffusion decoder with a VAE encoder for unified representation, producing virtual-satellite views supervised by real satellite targets.
  \textbf{(d) Comparison.} On University-1652, the proposed method occupies the top-right region of the accuracy plot, indicating competitive, often superior, cross-view retrieval performance for the satellite to drone task.}
  \label{Motivation}
\end{figure*}

Accurate cross-view matching faces two core challenges. As shown in \hyperref[Motivation]{Figure 1(a)}, the first is the disparity in imaging geometry and platform: UAVs typically fly at low altitudes with pronounced oblique look angles, whereas satellites acquire imagery from near-nadir, orthographic perspectives. This yields marked inconsistencies in scale, viewpoint, and occlusion patterns. The second is cross-temporal domain shift: seasonal changes alter the appearance of vegetation and water bodies; day/night and weather variations affect illumination; and dynamic objects such as vehicles appear or disappear, all of which further exacerbate the visual domain gap.

A large body of effective methods has sought to address the above challenges and can be broadly grouped, from a representation-learning perspective, into two families. Discriminative approaches~\cite{r9,r10,r11} employ metric learning to pull UAV and satellite images closer in a high-dimensional feature space, enabling efficient retrieval. Generative approaches~\cite{r12,r13,r14} attempt to synthesize pseudo-satellite images to narrow the domain gap. However, both lines of work typically rely on paired or annotated data: discriminative models degrade when transferred to new domains and often demand costly re-annotation, further constrained by regional compliance and data availability; generative models still struggle with cross-domain stability and controllability. GANs are notoriously unstable to train~\cite{r15}, and diffusion models~\cite{r16} in UAV localization often lack effective text conditioning and structural constraints. These issues are illustrated in \hyperref[Motivation]{Figure 1(b)}.

To reduce reliance on manual annotations, several recent efforts have emerged. \cite{r17} employs pseudo-labels and cross-view alignment to enable self-supervised training, achieving competitive results without labels; however, pseudo-labels can become noisy in complex environments, which limits the ultimate performance. \cite{r18} constructs 3D neural fields and optimizes them iteratively, markedly improving cross-view accuracy without paired data, yet the 3D reconstruction depends on an initial COLMAP~\cite{r19} pipeline that can fail in sparsely textured scenes.

To address these challenges, we propose a method that couples 3D reconstruction with image inpainting (see \hyperref[Motivation]{Figure 1(c)}). High-quality UAV orthophotos are rendered via projective geometry or 3D Gaussian Splatting, and a text-free diffusion model, guided by structural priors, performs UAV to satellite view synthesis and feature extraction. Departing from the conventional synthesize pseudo imagery then retrieve paradigm, our inference stage uses the high-dimensional representations produced by the diffusion model’s visual autoencoder (VAE) directly for retrieval, thereby preserving structural consistency during synthesis while achieving efficient and accurate matching.

The main contributions of this paper are as follows:
\begin{itemize}
  \item We propose a zero-training UAV to near-nadir pseudo-satellite rendering pipeline featuring orthographic projection, layered roof/ground rasterization, inpainting-based hole filling and style harmonization, and robustness to large look angles and sparse textures; when COLMAP fails, a perspective-projection fallback preserves engineering practicality.
  \item We employ a text-free conditional diffusion model: text prompts are replaced by visual prompts, and edge/semantic/depth priors are injected across diffusion timesteps via ControlNet (with LoRA), yielding multi-scale structural representations that are insensitive to viewpoint changes.
  \item We introduce a multi-scale, translation-invariant wavelet loss with uncertainty weighting, which explicitly aligns multi-band textures and edge details while adaptively balancing competing objectives.
  \item We design a single-forward latent-space retrieval scheme: dispensing with iterative denoising and image decoding, we directly use the conditioned latent as the UAV query descriptor and the VAE posterior mean as the satellite descriptor in the same latent space; retrieval is via cosine similarity.
\end{itemize}

\section{Related Work}

\textbf{Cross-view generative geolocalization}: Early studies~\cite{r12} predominantly adopted generative adversarial networks (GANs) to translate ground-level imagery into aerial/satellite views, thereby narrowing the viewpoint gap and assisting retrieval-based matching. Subsequently, \cite{r13,r30} progressively introduced geometric constraints and leveraged explicit geometric transformations together with feature fusion to improve cross-domain robustness. In recent years, diffusion models—by virtue of higher generative fidelity and training stability—have become the main line for cross-view synthesis and localization~\cite{r14,r21,r22}: on one front, they enable joint restoration and representation optimization under adverse weather and rapidly time-varying scenes~\cite{r23}; on another, by combining structured layouts (e.g., bird’s-eye view/BEV, point–line primitives) with multimodal fusion, they steadily enhance both synthesizability and matchability across UAV–satellite and ground–aerial viewpoints~\cite{r24}. Meanwhile, some works~\cite{r25} have begun to explore the discriminative potential of text-free diffusion, opening a path toward unified generative–discriminative representations. Given that existing visual localization datasets~\cite{r26} contain only a small amount of reliable text annotation, we do not follow the route of augmenting with large-scale textual prompts; instead, we directly use UAV orthophotos reconstructed from 3D as image prompts to guide a text-free diffusion model for subsequent synthesis and feature extraction.

\textbf{UAV visual geolocalization}: The dataset introduced in~\cite{r5} was the first to systematically incorporate UAV-view geolocalization, supporting tasks such as UAV-view target localization and navigation, and establishing a benchmark for cross-view retrieval. Along the axes of representation learning and matching, methods have evolved in terms of orientation encoding, locality, partitioning, alignment, generalization, and multimodality. Specifically, \cite{r27} explicitly encodes pixel-wise orientation between ground panoramas and satellite images to improve azimuthal consistency in cross-view alignment; LPN~\cite{r28} employs a local pattern network with ring/square partitions to emphasize central context and strengthen local discriminability; FSRA~\cite{r29} aggregates multi-region features with a Transformer and performs region alignment to balance global and local cues; MBF~\cite{r30} fuses UAV state (altitude, camera pose) token embeddings with image embeddings and applies hierarchical bilinear pooling for robustness; SRLN~\cite{r31} performs Swin-based, orientation-guided multi-scale fusion to mitigate viewpoint inconsistency; CA-HRS~\cite{r32} learns content-aware hierarchical selection factors to focus on salient regions; SDPL~\cite{r33} introduces shift-dense partitioning and shift fusion to suppress degradation from positional bias and scale variation; MFRGN~\cite{r34} integrates multi-scale and global–local features to improve cross-domain generalization; TirSA~\cite{r35} adopts a three-stage pipeline of self-supervised feature enhancement, adaptive feature integration, and post-processing; MCCG~\cite{r36} leverages ConvNeXt-based multi-classifier complementarity to obtain diverse discriminative features; MEAN~\cite{r37} performs multi-level progressive enhancement and cross-domain alignment to emphasize consistent/invariant representations; QDFL~\cite{r38} mines viewpoint-invariant vectors via adaptive query embeddings and feature-fusion units; and \cite{r39} jointly models center diffusion and edge radiation to balance key content with boundary details. Most related to our setting, CDM-Net~\cite{r24} uses a multimodal pipeline with conditional diffusion to synthesize near-orthographic views, thereby improving UAV-view localization. In contrast, we eschew text prompts and standard text-conditioned diffusion, and instead treat rendered orthophotos as strong-prior image prompts, which better align with the data availability and structural characteristics of UAV scenarios.

\section{Methods}
This section details our diffusion-based UAV visual geolocalization pipeline. We first describe the synthesis of UAV orthophotos that serve as paired image prompts. We then introduce a text-free, structure-guided diffusion model that translates oblique UAV views into satellite-style imagery to bridge the large viewpoint gap. Next, we extract descriptors from the model’s built-in VAE encoder for both UAV and satellite views to enable efficient, accurate retrieval. Finally, we present the loss functions used to optimize the overall generation process.

\subsection{Geometric-Prior Orthophoto Rendering}
We construct a dataset of UAV–pseudo-satellite prompt pairs. For each oblique UAV image, an automated pipeline produces a near-nadir pseudo-satellite orthophoto as the subsequent diffusion model’s image prompt.

\begin{figure}
  \centering
  \includegraphics[width=0.49 \textwidth]{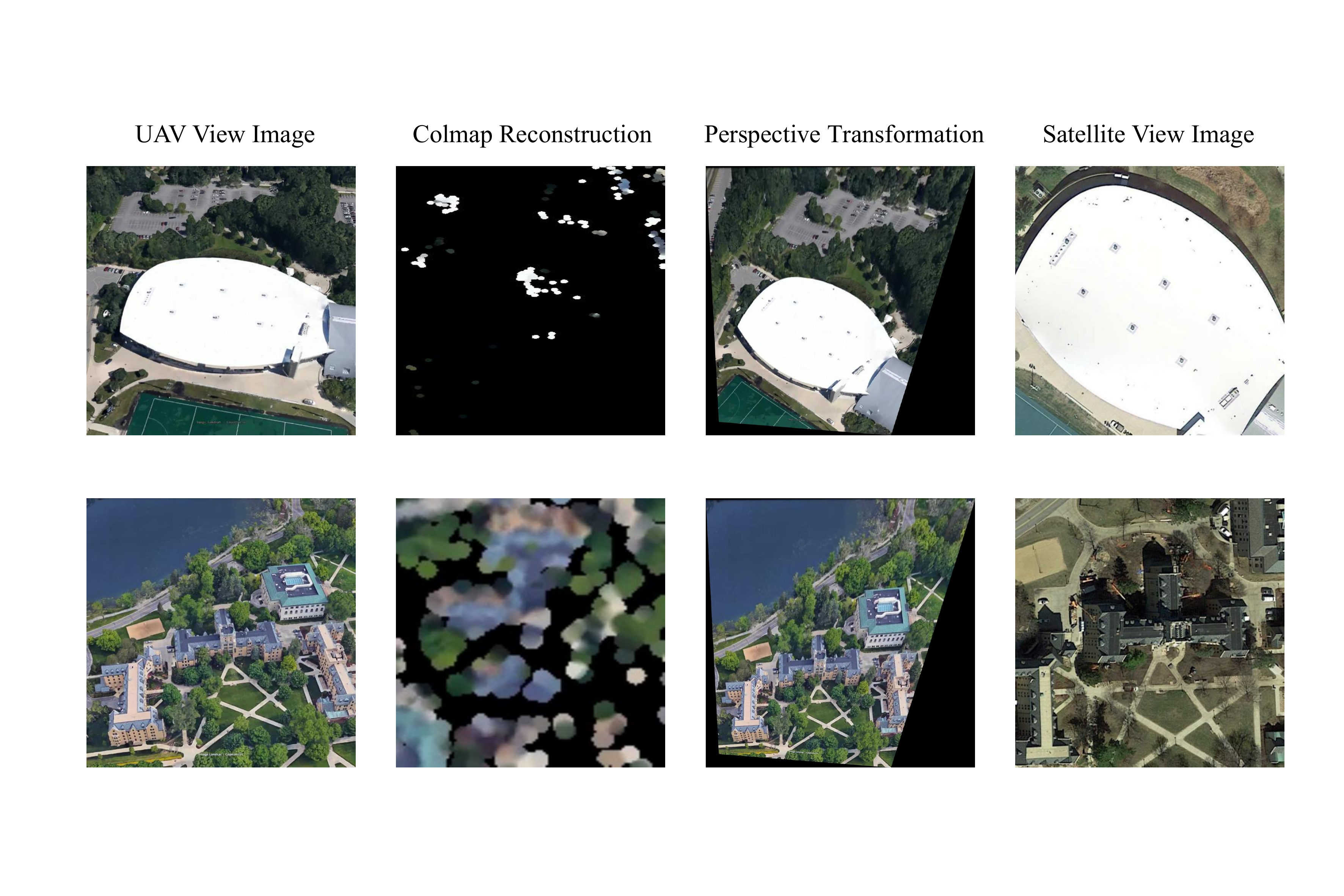}
  \caption{Traditional orthophoto rendering outcomes from left to right: the UAV viewpoint, failure of COLMAP reconstruction, consequences of perspective transformation and content loss, and the satellite image on the far right.}
  \label{failure_cases}
\end{figure}

Classical priors for converting oblique UAV imagery to orthophoto-like views rely on polar remapping~\cite{r12} or perspective transforms~\cite{r13}, which inevitably introduce stretching and distortions at large look angles or far from the principal axis. Recent 3D Gaussian Splatting (3DGS) offers strong novel-view synthesis but can fail when its COLMAP initialization breaks in sparsely textured scenes~\cite{r18}; it also tends to require per-scene training (e.g., scene-wise processing in University-1652), increasing engineering costs. Fig.~\ref{failure_cases} illustrates such failure cases. To this end, we keep COLMAP’s reconstruction unchanged and replace only the rendering stage with a lightweight 3D projection plus in-image inpainting, yielding realistic pseudo-satellite views in a zero-training, robust, and efficient manner.

Our pipeline first runs COLMAP to obtain sparse and dense reconstructions, producing a colored point cloud. Centered on orthographic projection in the ground reference frame, we combine a near-top roof-band fusion with a narrow-band ground averaging, and further apply in-image inpainting and mild style harmonization. We apply a fixed $20\%$ center crop to remove invalid black borders introduced by orthographic discretization or hole inpainting; this is a deterministic border-cleanup step (not an ablation). If COLMAP fails, we fall back to a perspective-based orthorectification to maintain practicality.

\subsubsection{Coordinate System and Orthographic Projection}
Given a 3D point set $\mathcal{P}=\{\mathbf{x}_i\in\mathbb{R}^3\}$, we estimate the ground plane via RANSAC:
\begin{equation}
\pi:\; a x + b y + c z + d = 0,\qquad 
\mathbf{n}=\frac{(a,b,c)}{\|(a,b,c)\|_2}.
\label{eq:plane}
\end{equation}
Construct an orthonormal basis $\{\mathbf{u},\mathbf{v},\mathbf{n}\}$ with $\mathbf{n}$ as the plane normal. Let $\bar{\mathbf{p}}$ be the centroid of the points used for plane fitting. Each point is expressed in plane coordinates
\begin{equation}
\mathbf{p}_i^{\pi}=
\begin{bmatrix}
u_i\\ v_i\\ h_i
\end{bmatrix},\quad
u_i=\langle \mathbf{x}_i-\bar{\mathbf{p}},\,\mathbf{u}\rangle,\;
v_i=\langle \mathbf{x}_i-\bar{\mathbf{p}},\,\mathbf{v}\rangle,\;
h_i=\langle \mathbf{x}_i-\bar{\mathbf{p}},\,\mathbf{n}\rangle.
\label{eq:proj}
\end{equation}
If the height distribution indicates a downward normal, we flip the sign to make upward positive. On the image plane, with pixel scale $r$ (m/px) and bounding-box origin $(u_{\min},v_{\min})$, the orthographic discretization is
\begin{equation}
x_i=\Big\lfloor \tfrac{u_i-u_{\min}}{r}\Big\rfloor,\qquad
y_i=H-1-\Big\lfloor \tfrac{v_i-v_{\min}}{r}\Big\rfloor.
\label{eq:pix}
\end{equation}

\subsubsection{Adaptive Resolution and SSAA}
To balance quality and memory, we estimate the ground coverage area $A$ and point count $N$ within a small-$|h|$ band. With target pixels $\widehat{P}\approx N/\rho$ ($\rho$: desired points per pixel), we set
\begin{equation}
r = \operatorname{clip}\!\left(\sqrt{\frac{A}{\widehat{P}}},\, r_{\min},\, r_{\max}\right),
\qquad \text{s.t. } H\times W \le P_{\max}.
\label{eq:res}
\end{equation}
We render on a supersampled grid (SSAA) and downsample to the base resolution via Lanczos to suppress aliasing.

\subsubsection{Layered Rasterization}
Roofs occupy the upper envelope in top-down height, while the ground clusters near $h\!\approx\!0$. We therefore adopt a two-band layering.

\noindent\emph{Roof layer.} For pixel $p$, compute the local maximum height $h_{\max}(p)$. Fuse only within the top band $[h_{\max}(p)-\Delta,\;h_{\max}(p)]$ with weights
\begin{equation}
w_i=\exp\!\left(\frac{h_i-h_{\max}(p)}{\tau}\right),\qquad \tau\approx\frac{\Delta}{2},
\label{eq:roofw}
\end{equation}
and accumulate samples via a small Gaussian disc splat (radius $\approx2$\,px) to obtain color $c^{\text{roof}}(p)$ and occupancy $\alpha^{\text{roof}}(p)$. If the hit count is below $m_{\min}$, we clear the pixel to avoid single-point artifacts.

\noindent\emph{Ground layer.} Average within a narrow ground band $[-b,+b]$ (we use $b=0.12$), again with small-radius splats, to yield $c^{\text{gnd}}(p)$. The final composite is
\begin{equation}
c(p)=\alpha^{\text{roof}}(p)\,c^{\text{roof}}(p)+\big(1-\alpha^{\text{roof}}(p)\big)\,c^{\text{gnd}}(p).
\label{eq:composite}
\end{equation}

\subsubsection{In-Image Inpainting}
To fill holes due to missing samples, we apply a state-of-the-art image inpainting method~\cite{r71}. The rendered pseudo-satellite image serves as the visual prompt for the second-stage diffusion model.

\subsection{Text-Free Controlled Cross-View Synthesis}
Given the scarcity and unreliability of textual prompts in UAV localization datasets~\cite{r3,r4,r6,r7}, and the unpredictability of prompt engineering, we adopt a text-free conditioning scheme inspired by~\cite{r25}. Let $I_{\text{prompt}}$ be the rendered UAV orthophoto. A pretrained lightweight ResNet-18~\cite{r47} encoder $E_{\phi}$ extracts a global visual feature, which a learnable projector $P_{\psi}$ maps into the Stable Diffusion~\cite{r48} text-embedding space:
\begin{equation}
\mathbf{H}=P_{\psi}\big(E_{\phi}(I_{\text{prompt}})\big)\in\mathbb{R}^{77\times768}.
\label{eq:visprompt}
\end{equation}
Thus, the visual prompt $\mathbf{H}$ replaces textual tokens, leveraging diffusion priors without text, naturally matching an image-to-image paradigm in remote sensing.

\subsection{Structure-Guided Diffusion Feature Encoder}
Our goal is not the pixel output per se, but to harvest discriminative features formed under diffusion dynamics.

\subsubsection{Latent Space and Noise Injection}
Encode the UAV view $I_{d}$ via the VAE encoder to obtain $z_0$. During training, inject noise at timestep $t$:
\begin{equation}
z_t=\sqrt{\alpha_t}\,z_0+\sqrt{1-\alpha_t}\,\boldsymbol{\epsilon},\quad \boldsymbol{\epsilon}\sim\mathcal{N}(\mathbf{0},\mathbf{I}),
\label{eq:normdiff}
\end{equation}
forcing the encoder to learn robust multi-scale representations under diffusion perturbations.

At inference, we fix $t=t^\star$ (selected on the validation set); performance is stable for $t$ within a small neighborhood of $t^\star$.

\subsubsection{Multi-Modal Structural Conditioning}
From $I_{d}$ we extract a Canny edge map $M_c$~\cite{r50}, a semantic mask $M_s$~\cite{r51}, and a relative depth map $M_d$~\cite{r52}. We concatenate them into
\begin{equation}
\mathcal{M}=[M_c,\,M_s,\,M_d]\in\mathbb{R}^{B\times3\times H\times W},
\label{eq:masks}
\end{equation}
and fuse them via a light conditional encoder $g_{\theta}$ to produce the ControlNet condition:
\begin{equation}
\mathbf{C}=g_{\theta}(\mathcal{M})\in\mathbb{R}^{B\times3\times H\times W}.
\label{eq:cond}
\end{equation}
This visible geometry enforces spatial layout consistency with the input UAV image.

\subsubsection{Diffusion-Aware Feature Extraction}
We freeze the ControlNet backbone and keep only the LoRA~\cite{r58} adapters trainable. Given a noise latent $z_t$, a diffusion time step $t$, a visual prompt $\mathbf{H}$, and a structural condition $\mathbf{C}$, a single forward pass (without full sampling) produces multi-scale downsampled features $\{f^{(k)}\}_{k=1}^{K}$ and a mid-level feature $f^{\text{(mid)}}$:
\begin{equation}
\{f^{(k)}\}_{k=1}^{K},\; f^{\text{(mid)}} \;=\; \mathrm{ControlNet}\!\left(z_t,\, t,\, \mathbf{H},\, \mathbf{C}\right).
\label{eq:features}
\end{equation}
Formed under joint generative constraints, these features are robust to viewpoint changes and sensitive to semantic and geometric structure.

\paragraph*{Descriptor Formation (train/inference readouts).}
From the mid-level representation, a lightweight latent head $R_{\omega}$ predicts a conditioned latent in the satellite domain that lives in the Stable Diffusion VAE latent space:
\begin{equation}
\hat{\mathbf{z}}_{s} \;=\; R_{\omega}\!\left(f^{\text{(mid)}}\right).
\label{eq:latent}
\end{equation}
For each satellite image $I_{s}$, the diffusion model's VAE encoder produces the posterior mean,
\begin{equation}
\boldsymbol{\mu}_{s} \;=\; \mathrm{Enc}_{\mathrm{VAE}}\!\left(I_{s}\right).
\label{eq:vae_sat}
\end{equation}
We then obtain retrieval descriptors by applying a readout operator $\varphi(\cdot)$ (default: \emph{flatten}) followed by $\ell_2$ normalization:
\begin{equation}
\mathbf{q}_{d} = \mathrm{norm}\!\big(\varphi(\hat{\mathbf{z}}_{s})\big), \qquad
\mathbf{r}_{s} = \mathrm{norm}\!\big(\varphi(\boldsymbol{\mu}_{s})\big),
\label{eq:readouts}
\end{equation}
and compute cosine similarity,
\begin{equation}
\mathrm{sim}\!\left(I_{d}, I_{s}\right)=\cos\!\left(\mathbf{q}_{d},\, \mathbf{r}_{s}\right).
\label{eq:sim}
\end{equation}
During training, $R_{\omega}$ is optimized jointly with the rest of the network; at inference, we keep $R_{\omega}$ and the VAE encoder for descriptor extraction while disabling the pixel decoder (Sec.~\ref{sec:pixeldecoder}).

\subsection{Structure-Aware Pixel Decoder (Training Only)}
\label{sec:pixeldecoder}
To train the generator, a U-Net-style \emph{pixel decoder} $D_{\text{img}}$ upsamples and fuses multi-scale features (with skip connections from ControlNet), ConvNeXt modules, residual blocks, and CBAM attention~\cite{r54}, finally producing a satellite orthophoto $\hat{I}_s$ via a convolution and $\tanh$. This branch provides pixel-level supervision for the losses in Sec.~\ref{sec:losses}. 
At inference, we disable $D_{\text{img}}$ entirely; retrieval uses only VAE-latent readouts (flatten $+$ $\ell_2$) from Eq.~\eqref{eq:readouts}, without iterative denoising or pixel-space decoding.

\subsection{Multi-Objective Losses}
\label{sec:losses}
We jointly optimize perceptual quality, fine textures, and structural consistency via uncertainty-weighted multi-task losses.

\subsubsection{Perceptual Loss}
Using VGG-16~\cite{r55} features $\phi(\cdot)$, we align high-level semantics/style:
\begin{equation}
\mathcal{L}_{\mathrm{perc}}=\big\|\phi(\hat{I}_s)-\phi(I_s)\big\|_{1}.
\label{eq:perc}
\end{equation}

\subsubsection{Weighted Shift-Invariant Wavelet Detail Loss}
Traditional pixel-space losses (e.g., MSE or vanilla $\ell_1$) emphasize global brightness/color matching but are weak at penalizing mismatches in \emph{texture granularity} and \emph{edge sharpness}. Two images may look similar under such losses while one is noticeably blurrier to a human observer. Classical discrete wavelet transforms (DWT) better capture details but introduce \emph{downsampling} at each level, making detail coefficients highly \emph{shift-sensitive}: a small spatial offset (e.g., a building edge at $(10,10)$ in $I_s$ versus $(11,11)$ in $\hat{I}_s$) can yield disproportionately large coefficient differences unrelated to perceived quality.

To overcome this, we adopt the \emph{stationary wavelet transform} (SWT, à trous), which removes decimation and keeps the spatial resolution \emph{unchanged} at all scales. Consequently, detail coefficients at each level are \emph{aligned pixelwise} with the input, substantially improving \emph{shift invariance} and allowing us to compare “the same location’s details’’ between $\hat{I}_s$ and $I_s$. In practice, we apply SWT per channel and per scale, so that fine-scale coefficients remain sensitive to thin, high-frequency edges while coarser scales summarize broader textural patterns.

Let $D_j(\cdot)$ denote the SWT detail coefficients at levels $j=1,\dots,L$. Lower $j$ emphasizes high-frequency structures (e.g., building boundaries, road markings) that most strongly affect perceived sharpness, whereas higher $j$ captures coarser textures and larger structural motifs. We therefore aggregate per-scale discrepancies with nonnegative weights $w_j$ to reflect their different perceptual/structural importance in UAV geolocalization (e.g., placing more emphasis on $j=1$–$2$ for crisp edges while retaining nonzero weights at larger $j$ to preserve layout cues). Using an $\ell_1$ penalty promotes sparsity in the residuals and is empirically more faithful to edge fidelity than $\ell_2$ in our setting. The final loss reads:
\begin{equation}
\mathcal{L}_{\mathrm{swt}}=\sum_{j=1}^{L} w_j\,\big\|D_j(\hat{I}_s)-D_j(I_s)\big\|_{1}.
\label{eq:swt}
\end{equation}
This design jointly achieves (i) \emph{detail awareness} via multi-scale wavelet coefficients, (ii) \emph{shift invariance} via SWT’s undecimated filters, and (iii) \emph{task adaptivity} via learnable or preset $\{w_j\}$, yielding sharper edges and more faithful textures without sacrificing robustness to slight misalignment.

\subsubsection{Structural Mask Loss}
Focusing optimization on geometry/semantics-sensitive regions (derived from the same modalities as $\mathcal{M}$), with masks $\mathcal{S}=\{S_m\}_{m=1}^{M}$ and weights $\lambda_m$:
\begin{equation}
\mathcal{L}_{\mathrm{mask}}=\sum_{m=1}^{M}\lambda_m\,\big\|\big(\hat{I}_s-I_s\big)\odot S_m\big\|_{1}.
\label{eq:mask}
\end{equation}
(We use $\mathcal{S}$ to avoid conflict with the conditioning tensor $\mathcal{M}$ in Eq.~\eqref{eq:masks}.)

\subsubsection{Uncertainty-Based Weighting}
We combine losses via learned task uncertainties:
\begin{equation}
\mathcal{L}_{\mathrm{total}}=\sum_{i}\exp(-s_i)\,\mathcal{L}_i + s_i,\quad 
\mathcal{L}_i\in\{\mathcal{L}_{\mathrm{perc}},\mathcal{L}_{\mathrm{swt}},\mathcal{L}_{\mathrm{mask}}\},
\label{eq:uncertainty}
\end{equation}
where $s_i=\log\sigma_i^2$ are learnable.

\subsection{Feature Extraction for Localization}
The first-stage generator implicitly aligns cross-domain geometry and layout. For inference-time localization, we reuse the unified VAE latent space without pixel-space decoding: given an oblique UAV input $I_d$ and condition $(\mathbf{H},\mathbf{C})$, a single ControlNet forward pass (with $t=t^\star$) produces $\hat{\mathbf{z}}_s$; for a reference satellite image $I_s$, the VAE encoder yields $\boldsymbol{\mu}_s$. Descriptors are computed via \emph{flatten} readout and $\ell_2$ normalization as in Eq.~\eqref{eq:readouts}, and retrieval uses cosine similarity (Eq.~\ref{eq:sim}). This design requires no iterative denoising and no additional training beyond the first-stage generator.

\section{Experiments}

\begin{table}[t]
  \centering
  % University-1652 table caption
  \caption{Results on University-1652. We report Recall@1 and AP (in \%). The arrow $\uparrow$ indicates higher is better. Bold indicates the best result in each column.}
  \label{tab:uni1652}
  \footnotesize
  \begin{tabular}{lcccc}
    \toprule
    \multirow{2}{*}{Methods} &
    \multicolumn{2}{c}{\textbf{drone to satellite}} &
    \multicolumn{2}{c}{\textbf{satellite to drone}} \\
    \cmidrule(lr){2-3}\cmidrule(lr){4-5}
    & Recall@1$\uparrow$ & AP$\uparrow$ & Recall@1$\uparrow$ & AP$\uparrow$ \\
    \midrule
    Instance Loss~\cite{r70} & 58.23 & 62.91 & 74.47 & 59.45 \\
    LCM~\cite{r59}           & 66.65 & 70.82 & 79.89 & 65.38 \\
    LPN~\cite{r28}           & 75.93 & 79.14 & 86.45 & 74.79 \\
    FSRA~\cite{r29}          & 84.51 & 86.71 & 88.45 & 83.37 \\
    IFSs~\cite{r60}          & 86.06 & 88.08 & 91.44 & 85.73 \\
    MBF~\cite{r30}           & 89.05 & 90.61 & 93.15 & 88.17 \\
    SeGCN~\cite{r61}         & 89.18 & 90.89 & 94.29 & 89.65 \\
    MCCG~\cite{r36}          & 89.64 & 91.32 & 94.30 & 89.39 \\
    SDPL~\cite{r33}          & 90.16 & 91.64 & 93.58 & 89.45 \\
    MFJR~\cite{r62}          & 91.87 & 93.15 & 95.29 & 91.51 \\
    CCR~\cite{r63}           & 92.54 & 93.78 & 95.15 & 91.80 \\
    Sample4Geo~\cite{r64}    & 92.65 & 93.81 & 95.14 & 91.39 \\
    SRLN~\cite{r31}          & 92.70 & 93.77 & 95.14 & 91.97 \\
    JRN-Geo~\cite{r39}       & 92.79 & 93.91 & 95.98 & 92.17 \\
    MEAN~\cite{r37}          & 93.55 & 94.53 & 96.01 & 92.08 \\
    LRFR~\cite{r65}          & 94.14 & 95.09 & 95.72 & 93.22 \\
    DAC~\cite{r66}           & 94.67 & 95.50 & 96.43 & 93.79 \\
    QDFL~\cite{r38}          & 95.00 & 95.83 & 97.15 & 94.57 \\
    CDM-Net~\cite{r24}       & 95.13 & 96.04 & 96.68 & 94.05 \\
    CGSI~\cite{r67}          & \textbf{95.45} & \textbf{96.10} & 96.58 & 95.38 \\
    \textbf{Ours}            & 94.10 & 95.93 & \textbf{98.14} & \textbf{97.83} \\
    \bottomrule
  \end{tabular}
\end{table}

\begin{table*}[t]
  \centering
  \caption{SUES-200 by flight altitude. For each altitude, we report Recall@1 and AP (in \%) in both directions. The arrow $\uparrow$ indicates higher is better. Bold indicates the best result in each column.}

  \label{tab:sues_all_in_one}
  \scriptsize
  \setlength{\tabcolsep}{2.8pt}
  \resizebox{\textwidth}{!}{%
  \begin{tabular}{l*{16}{c}}
    \toprule
    \multirow{3}{*}{Method} &
      \multicolumn{4}{c}{\textbf{150 m}} &
      \multicolumn{4}{c}{\textbf{200 m}} &
      \multicolumn{4}{c}{\textbf{250 m}} &
      \multicolumn{4}{c}{\textbf{300 m}} \\
    \cmidrule(lr){2-5}\cmidrule(lr){6-9}\cmidrule(lr){10-13}\cmidrule(lr){14-17}
    & \multicolumn{2}{c}{\textbf{drone to satellite}} & \multicolumn{2}{c}{\textbf{satellite to drone}}
    & \multicolumn{2}{c}{\textbf{drone to satellite}} & \multicolumn{2}{c}{\textbf{satellite to drone}}
    & \multicolumn{2}{c}{\textbf{drone to satellite}} & \multicolumn{2}{c}{\textbf{satellite to drone}}
    & \multicolumn{2}{c}{\textbf{drone to satellite}} & \multicolumn{2}{c}{\textbf{satellite to drone}} \\
    \cmidrule(lr){2-3}\cmidrule(lr){4-5}
    \cmidrule(lr){6-7}\cmidrule(lr){8-9}
    \cmidrule(lr){10-11}\cmidrule(lr){12-13}
    \cmidrule(lr){14-15}\cmidrule(lr){16-17}
    & Recall@1$\uparrow$ & AP$\uparrow$ & Recall@1$\uparrow$ & AP$\uparrow$
    & Recall@1$\uparrow$ & AP$\uparrow$ & Recall@1$\uparrow$ & AP$\uparrow$
    & Recall@1$\uparrow$ & AP$\uparrow$ & Recall@1$\uparrow$ & AP$\uparrow$
    & Recall@1$\uparrow$ & AP$\uparrow$ & Recall@1$\uparrow$ & AP$\uparrow$ \\
    \midrule
    SUES Baseline~\cite{r6}        & 59.32 & 64.93 & 82.50 & 58.95 & 62.30 & 67.24 & 85.00 & 62.56 & 71.35 & 75.49 & 88.75 & 69.96 & 77.17 & 80.67 & 96.25 & 84.16 \\
    LCM~\cite{r59}        & 43.42 & 49.65 & 57.50 & 38.11 & 49.42 & 55.91 & 68.75 & 49.19 & 54.47 & 60.31 & 72.50 & 47.94 & 60.43 & 65.78 & 75.00 & 59.36 \\
    LPN~\cite{r28}        & 61.58 & 67.23 & 83.75 & 66.78 & 70.85 & 75.96 & 88.75 & 75.01 & 80.38 & 83.80 & 92.50 & 81.34 & 81.47 & 84.53 & 92.50 & 85.72 \\
    FSRA~\cite{r29}       & 68.25 & 73.45 & 83.75 & 76.67 & 83.00 & 85.99 & 90.00 & 85.34 & 90.68 & 92.27 & 93.75 & 90.17 & 91.95 & 91.95 & 95.00 & 92.03 \\
    IFSs ~\cite{r60}      & 77.57 & 81.30 & 93.75 & 79.49 & 89.50 & 91.40 & 97.50 & 90.52 & 92.58 & 94.21 & 97.50 & 96.03 & 97.40 & 97.92 & \textbf{100.00} & 97.66 \\
    MBF~\cite{r30}        & 85.62 & 88.21 & 88.75 & 84.74 & 87.43 & 90.02 & 91.25 & 89.95 & 90.65 & 92.53 & 93.75 & 90.65 & 92.12 & 93.63 & 96.25 & 91.60 \\
    SeGCN~\cite{r61}      & 90.80 & 92.32 & 93.75 & 92.45 & 91.93 & 93.41 & 95.00 & 93.65 & 92.53 & 93.90 & 96.25 & 94.39 & 93.33 & 94.61 & 97.50 & 94.55 \\
    MCCG~\cite{r36}       & 82.22 & 85.47 & 93.75 & 89.72 & 89.38 & 91.41 & 93.75 & 92.21 & 93.82 & 95.04 & 96.25 & 96.14 & 95.07 & 96.20 & 98.75 & 96.64 \\
    SDPL~\cite{r33}       & 82.95 & 85.82 & 93.75 & 83.75 & 92.73 & 94.07 & 96.25 & 92.42 & 96.05 & 96.69 & 97.50 & 95.65 & 97.83 & 98.05 & 96.25 & 96.17 \\
    MFJR~\cite{r62}       & 88.95 & 91.05 & 95.00 & 89.31 & 93.60 & 94.72 & 96.25 & 94.69 & 95.42 & 96.28 & 97.50 & 96.92 & 97.45 & 97.84 & 98.75 & 97.14 \\
    CCR~\cite{r63}        & 87.08 & 89.55 & 92.50 & 88.54 & 93.57 & 94.90 & 97.50 & 95.22 & 95.42 & 96.28 & 97.50 & 97.10 & 96.82 & 97.39 & 97.50 & 97.49 \\
    Sample4Geo~\cite{r64} & 92.60 & 94.00 & 97.50 & 93.63 & 97.38 & 97.81 & 98.75 & 96.70 & 98.28 & 98.64 & 98.75 & 98.28 & 99.18 & 99.36 & 98.75 & 98.05 \\
    SRLN~\cite{r31}       & 89.90 & 91.90 & 93.75 & 93.01 & 94.32 & 95.65 & 97.50 & 95.08 & 95.92 & 96.79 & 97.50 & 96.52 & 96.37 & 97.21 & 97.50 & 96.71 \\
    JRN-Geo~\cite{r39}    & 85.30 & 87.58 & 93.75 & 86.93 & 93.23 & 94.66 & 97.75 & 93.12 & 94.47 & 97.28 & 98.75 & 96.81 & 97.50 & 98.09 & 98.75 & 97.20 \\
    MEAN~\cite{r37}       & 95.50 & 96.46 & 97.50 & 94.75 & \textbf{98.38} & \textbf{98.72} & \textbf{100.00} & 97.09 & 98.95 & 99.17 & \textbf{100.00} & 98.28 & \textbf{99.52} & \textbf{99.63} & \textbf{100.00} & \textbf{99.21} \\
    LRFR~\cite{r65}       & 92.87 & 94.49 & 96.67 & 92.62 & 97.03 & 97.77 & 96.67 & 95.28 & 97.87 & 98.38 & 98.33 & 96.05 & 98.30 & 98.73 & 98.33 & 96.67 \\
    DAC~\cite{r66}        & \textbf{96.80} & \textbf{97.54} & 97.50 & 94.06 & 97.48 & 97.97 & 98.75 & 96.66 & 98.20 & 98.62 & 98.75 & 98.09 & 97.58 & 98.14 & 98.75 & 98.87 \\
    QDFL~\cite{r38}       & 93.97 & 95.42 & \textbf{98.75} & 95.10 & 98.25 & 98.67 & 98.75 & 97.92 & \textbf{99.30} & \textbf{99.48} & \textbf{100.00} & \textbf{99.07} & 99.31 & 99.48 & \textbf{100.00} & 99.07 \\
    CDM-Net~\cite{r24}    & 93.78 & 95.16 & 95.25 & 92.24 & 97.62 & 98.16 & 98.50 & 96.40 & 98.28 & 98.69 & 99.00 & 97.60 & 99.20 & 99.31 & 99.00 & 98.01 \\
    CGSI~\cite{r67}       & 95.95 & 96.80 & 97.50 & 96.22 & 97.72 & 98.15 & 98.75 & 97.62 & 97.60 & 98.03 & 98.75 & 98.01 & 97.83 & 98.23 & 98.75 & 97.92 \\
    \textbf{Ours}         & 92.90 & 93.93 & 97.50 & \textbf{97.67} & 93.67 & 94.66 & 97.50 & \textbf{98.13} & 93.87 & 94.71 & 97.50 & 97.69 & 93.55 & 94.49 & 98.75 & 97.63 \\
    \bottomrule
  \end{tabular}}
  \vspace{2pt}
\end{table*}

\subsection{Implementation Details}
\textbf{Training.}
We train four components end-to-end: the projector $P_{\psi}$, the conditional encoder $g_{\theta}$, the latent head $R_{\omega}$ (for VAE-space alignment), and the pixel decoder $D_{\text{img}}$ (for image supervision). The ControlNet backbone is kept frozen, and we fine-tune only LoRA~\cite{r58} adapters to reduce trainable parameters and memory while preserving backbone stability, making the structural pathway more sensitive to cross-view geometry. The objective is a perception–texture–structure joint loss in Sec.~\ref{sec:losses}: a perceptual term aligns high-level semantics, a texture term constrains mid-frequency details, and a structure term enforces multi-scale (wavelet-domain) edge and contour consistency. Uncertainty-based weighting adaptively balances these terms.

\textbf{Inference (Localization).}
Given $(I_{d}, \mathcal{M})$, we build the condition $\mathbf{C}$, fix a time step $t=t^\star$, and run a single ControlNet forward pass to obtain the conditioned latent estimate $\hat{\mathbf{z}}_{s}$. The UAV \emph{query} and satellite \emph{reference} descriptors are computed by the \emph{flatten} readout with $\ell_2$ normalization as in Eq.~\eqref{eq:readouts} (for satellites we use the VAE posterior mean). Retrieval is performed via cosine similarity (Eq.~\ref{eq:sim}). No text input or iterative denoising is required, and the pixel decoder is disabled at test time.

\subsection{Datasets}
\textbf{University-1652}
 A multi-view, multi-source dataset for UAV geolocalization, containing images from UAV, satellite, and ground viewpoints. It covers 1{,}652 university buildings worldwide. Each building is associated with multiple UAV images captured from different angles and altitudes, helping models learn viewpoint-invariant features. To reduce the cost of real UAV flights, a portion of the UAV imagery is synthesized from Google Earth 3D models. The dataset also includes street-view and satellite imagery; the average number of images per location is about 58. The tasks include UAV-view target localization and UAV navigation, providing rich multi-view data for geolocalization and navigation.

\textbf{SUES-200.}
Designed for cross-view image matching with a focus on UAV localization, this dataset comprises 24{,}120 images from 200 sites, including UAV imagery captured at four altitudes (150 m, 200 m, 250 m, and 300 m) and the corresponding satellite images. It spans diverse scenes—parks, schools, lakes, and public buildings—facilitating the learning of discriminative features across environments. SUES-200 explicitly emphasizes the impact of flight altitude on matching performance and supports two primary tasks: UAV-view target localization and UAV navigation.

\subsection{Baselines}
We compare against advanced retrieval-based UAV localization methods.
Instance Loss~\cite{r70},
SUES Baseline~\cite{r6},
LCM~\cite{r59},
LPN~\cite{r28},
FSRA~\cite{r29},
IFSs~\cite{r60},
MBF~\cite{r30},
SeGCN~\cite{r61},
MCCG~\cite{r36},
SDPL~\cite{r33},
MFJR~\cite{r62},
CCR~\cite{r63},
Sample4Geo~\cite{r64},
SRLN~\cite{r31},
JRN-Geo~\cite{r39},
MEAN~\cite{r37},
LRFR~\cite{r65},
DAC~\cite{r66},
QDFL~\cite{r38},
CDM-Net~\cite{r24},
CGSI~\cite{r67}.

\subsection{Metrics}
We report Recall@1 ($\uparrow$) and AP (Average Precision, $\uparrow$) for two directions: \emph{drone to satellite} and \emph{satellite to drone}.

\begin{figure*}[t]
  \centering
  \includegraphics[width=\textwidth]{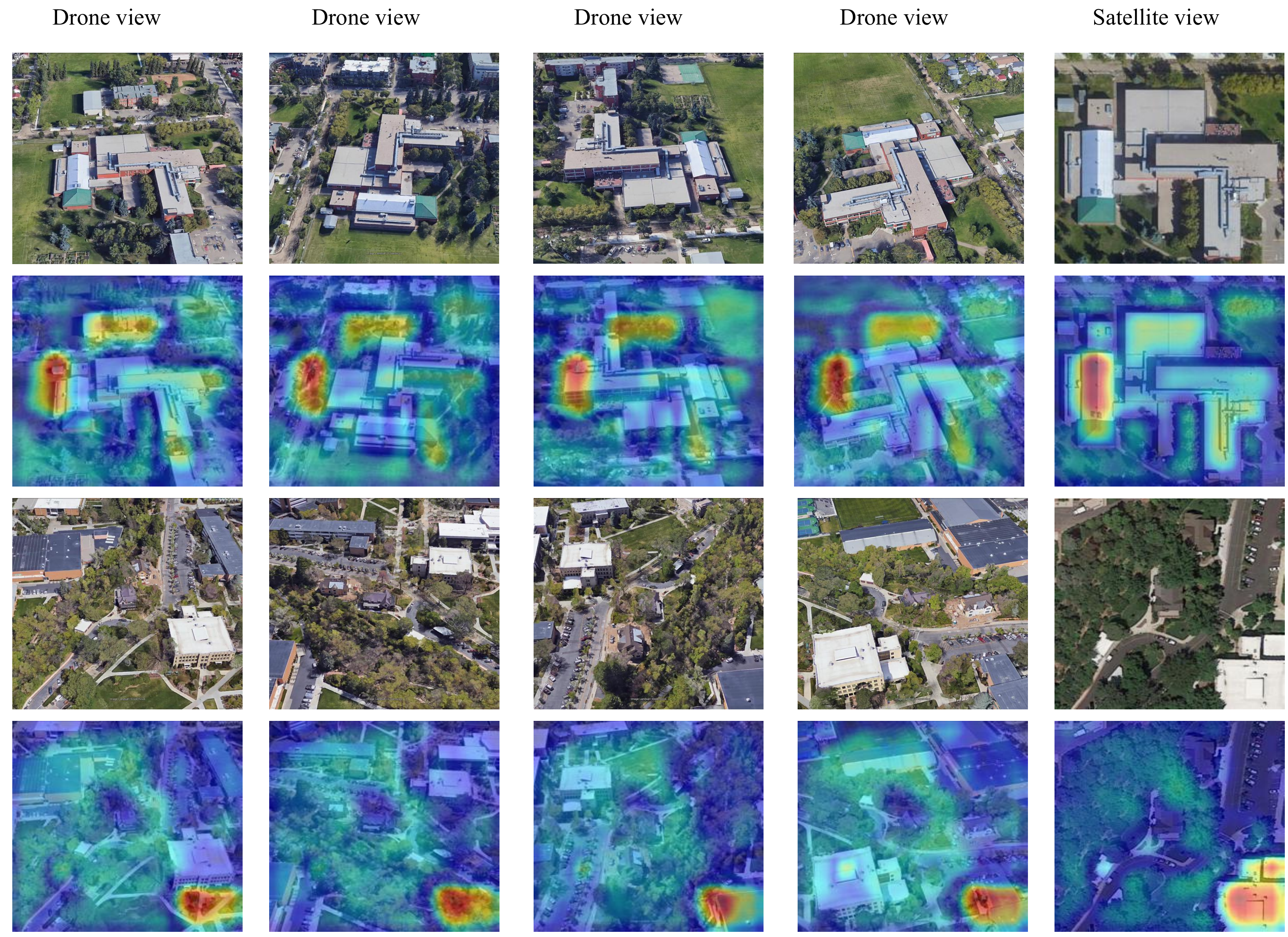}
  \caption{Qualitative response heatmaps on University-1652. For multiple drone viewpoints and the corresponding satellite reference, our method produces compact and spatially aligned activations that concentrate on geometry-consistent regions such as building footprints, long straight edges, and stable man-made structures. The hotspots remain co-located across-viewpoints and largely overlap with the satellite reference, indicating robust cross-view correspondence and reduced spurious responses. Overall, the heatmaps show tighter localization and better alignment between drone and satellite views after applying our approach.}
  \label{fig:heatmap}
\end{figure*}

\subsection{Quantitative Comparison}

\paragraph*{University-1652}
As shown in Table~\ref{tab:uni1652}, our method leads clearly in the \emph{satellite to drone} direction: Recall@1 is $98.14$, which is $+0.99$ over the second best QDFL, and AP is $97.83$, which is $+2.45$ over the second best CGSI. The directional average $(\text{R@1}+\text{AP})/2$ reaches $97.99$, well above CGSI’s $95.98$. In the \emph{drone to satellite} direction the gap to the top baseline is small: Recall@1 is $94.10$ versus $95.45$ for CGSI (difference $1.35$), and AP is $95.93$ versus $96.10$ for CGSI (difference $0.17$), giving a directional average of $95.02$ versus $95.78$ for CGSI. Macro averaging all four metrics yields $96.50$, which exceeds CGSI ($95.88$) and QDFL ($95.64$) by $+0.62$ and $+0.86$ respectively.

\paragraph*{SUES-200}
Table~\ref{tab:sues_all_in_one} provides a fine-grained view by flight altitude. Our advantage is most visible at lower altitudes. At $150$ m and $200$ m in the \emph{satellite to drone} direction, our AP values are $97.67$ and $98.13$, both column bests. At $150$ m our Recall@1 is $97.50$, slightly below QDFL’s $98.75$, but the higher AP indicates more stable ranking. When altitude increases to $250$ m and $300$ m, MEAN and QDFL reach extreme Recall@1 values, and our Recall@1 is slightly lower. Even so, our AP remains high at $97.69$ and $97.63$. Across altitudes in \emph{satellite to drone}, our Recall@1 values are $97.50$, $97.50$, $97.50$, and $98.75$ (variation $1.25$ points), and AP values are $97.67$, $98.13$, $97.69$, and $97.63$ (variation $0.50$ points), showing strong robustness to height changes. In the \emph{drone to satellite} direction, Recall@1 values center around $93.5$ and AP values around $94.5$, with variations below $1$ point, which also indicates consistent behavior across heights.

\paragraph*{Takeaways}
(1) At lower altitudes where viewpoint differences are larger, our method delivers the strongest AP in the \emph{satellite to drone} direction and gives more reliable rankings on difficult queries.
(2) Performance is smoother across altitudes, especially the \emph{satellite to drone} AP, which is nearly insensitive to height changes. This stability is favorable for real flights and direct deployment.
(3) On University-1652 we achieve a higher macro average than strong baselines, indicating balanced strength across directions and metrics rather than excelling in only one case.

\paragraph*{Analysis}
Our advantages stem from four components. First, \emph{source level structural alignment}: rendered UAV orthophotos used as image conditions explicitly align geometry, reduce pitch and perspective discrepancies, and boost AP in the \emph{satellite to drone} direction. Second, \emph{ControlNet with LoRA without text prompts}: by avoiding scarce or noisy text and tuning only a small set of parameters, the structural pathway focuses on cross-view and multi-scale geometric priors while the backbone remains stable. Third, \emph{VAE global representations for retrieval}: the method removes the synthesize then embed pipeline and produces compact, searchable descriptors in a single forward pass, which reduces latency and error accumulation. Fourth, a \emph{multi- scale wavelet loss}: it enforces edge and texture consistency at multiple scales and improves ranking quality, especially at low altitude and in scenes with complex structures. Overall, the combination of structural priors and efficient descriptor generation yields stronger stability at low altitude, leading AP in the \emph{satellite to drone} direction, and a superior macro average on University-1652.

\subsection{Ablation Study}
\label{sec:ablation}

We conduct an ablation study on University-1652 along four variants: (i) perspective-only projection, (ii) using text prompts instead of text-free conditioning, (iii) removing the multi-scale wavelet loss, and (iv) the full model with all components enabled.

\begin{table}[t]
  \centering
  \caption{Ablation on University-1652 (\%, higher is better).}
  \label{tab:ablation_uni1652}
  \footnotesize
  \begin{tabular}{lcccc}
    \toprule
    \multirow{2}{*}{Variant} &
    \multicolumn{2}{c}{\textbf{drone to satellite}} &
    \multicolumn{2}{c}{\textbf{satellite to drone}} \\
    \cmidrule(lr){2-3}\cmidrule(lr){4-5}
    & Recall@1$\uparrow$ & AP$\uparrow$ & Recall@1$\uparrow$ & AP$\uparrow$ \\
    \midrule
    Perspective-only projection        & 85.32 & 87.10  & 86.45 & 85.60 \\
    With text prompts                  & 93.71 & 95.45 & 97.21 & 97.15 \\
    Without wavelet loss               & 92.15 & 93.32 & 96.09 & 95.75 \\
    All components (ours)              & \textbf{94.10} & \textbf{95.93} & \textbf{98.14} & \textbf{97.83} \\
    \bottomrule
  \end{tabular}
\end{table}

\paragraph{Orthographic point-cloud projection vs. perspective-only.}
Perspective-only rendering leaves residual geometric misalignment under complex occlusions and large look angles. Incorporating orthographic point-cloud projection with transparency-based compositing strengthens orthorectified consistency, with more pronounced gains for the satellite to drone direction.

\paragraph{Effect of text-free conditioning.}
Removing text-related branches yields more stable training and faster convergence. On cross-domain data, it avoids prompt ambiguity and produces consistent improvements in Recall@1 and AP.

\paragraph{Multi-scale wavelet loss.}
Ablating this term leads to under-matching of high-frequency textures and a noticeable drop in AP. Restoring it improves AP for the satellite to drone direction and stabilizes the macro average.

\subsection{Heatmap Visualization}
\label{sec:heatmap}

We visualize response heatmaps to assess cross-view correspondence between drone and satellite imagery. After applying our approach, the activations become compact and spatially coherent, concentrating on geometry-consistent regions such as building footprints and long structural edges. The highlighted areas remain co-located across multiple drone viewpoints and closely overlap with the corresponding satellite reference, which indicates tighter localization, improved alignment, and a reduction of spurious responses. These observations support the claim that the proposed method strengthens cross-view consistency and yields more reliable retrieval cues.

\section{Limitations and Discussion}
\label{sec:limitations}

\noindent\textbf{Dependence on COLMAP.}
The pipeline relies on COLMAP to obtain reliable sparse and dense reconstructions. In scenes with extremely sparse textures or large water bodies, reconstruction may fail. We provide a perspective-based fallback and in-image hole filling to maintain functionality, but geometric consistency can degrade in such cases. Future work may investigate robust reconstruction under sparse or non-rigid settings, or replace the geometric initialization with learned depth priors that are less sensitive to low-texture regions.

\noindent\textbf{Quality of structural priors.}
Errors in semantic masks, depth estimates, and edge maps can propagate to latent-space alignment and affect retrieval quality. This limitation could be mitigated by more robust lightweight estimators, cross-modal consistency checks, and uncertainty-aware conditioning that down-weights unreliable priors during training and inference.

\noindent\textbf{Cross-domain robustness.}
Although the method improves viewpoint robustness, extreme seasonal changes, illumination shifts, and heavy occlusions may still induce residual appearance gaps. Domain generalization via style-robust augmentations, curriculum scheduling across seasons and cities, and feature normalization tailored to aerial–satellite mixtures are promising directions.

\section{Conclusion}
\label{sec:conclusion}

We propose \emph{DiffusionUavLoc}, a cross-view UAV localization framework that combines geometry-driven image prompting, text-free conditional diffusion, and unified retrieval in the Stable Diffusion VAE latent space. The system eliminates iterative denoising at inference and requires no text input, substantially improving practical applicability. On the University-1652 dataset, the method achieves leading performance in the satellite to drone direction and demonstrates strong overall competitiveness. Looking ahead, we plan to explore weakly supervised end-to-end training, improved generalization across cities and seasons, and the adaptive selection of diffusion timesteps and prior-quality weights to further enhance reliability in complex environments.

\bibliographystyle{IEEEtran} 
\bibliography{reference}

@ARTICLE{r1,
  author={Shirmohammadi, Shervin and Ferrero, Alessandro},
  journal={IEEE Instrumentation \& Measurement Magazine}, 
  title={Camera as the instrument: the rising trend of vision based measurement}, 
  year={2014},
  volume={17},
  number={3},
  pages={41-47},
  doi={10.1109/MIM.2014.6825388}}

@article{r2,
  title={Graph-based adaptive fusion of GNSS and VIO under intermittent GNSS-degraded environment},
  author={Gong, Zheng and Liu, Peilin and Wen, Fei and Ying, Rendong and Ji, Xingwu and Miao, Ruihang and Xue, Wuyang},
  journal={IEEE Transactions on Instrumentation and Measurement},
  volume={70},
  pages={1--16},
  year={2020},
  publisher={IEEE}
}

@article{r3,
  title={Leveraging map retrieval and alignment for robust UAV visual geo-localization},
  author={He, Mengfan and Liu, Jiacheng and Gu, Pengfei and Meng, Ziyang},
  journal={IEEE Transactions on Instrumentation and Measurement},
  volume={73},
  pages={1--13},
  year={2024},
  publisher={IEEE}
}

@article{r4,
  title={Exploring the best way for UAV visual localization under Low-altitude Multi-view Observation Condition: a Benchmark},
  author={Ye, Yibin and Teng, Xichao and Chen, Shuo and Li, Zhang and Liu, Leqi and Yu, Qifeng and Tan, Tao},
  journal={arXiv preprint arXiv:2503.10692},
  year={2025}
}

@inproceedings{r5,
  title={University-1652: A multi-view multi-source benchmark for drone-based geo-localization},
  author={Zheng, Zhedong and Wei, Yunchao and Yang, Yi},
  booktitle={Proceedings of the 28th ACM international conference on Multimedia},
  pages={1395--1403},
  year={2020}
}

@article{r6,
  title={SUES-200: A multi-height multi-scene cross-view image benchmark across drone and satellite},
  author={Zhu, Runzhe and Yin, Ling and Yang, Mingze and Wu, Fei and Yang, Yuncheng and Hu, Wenbo},
  journal={IEEE Transactions on Circuits and Systems for Video Technology},
  volume={33},
  number={9},
  pages={4825--4839},
  year={2023},
  publisher={IEEE}
}

@ARTICLE{r7,
  author={Dai, Ming and Zheng, Enhui and Feng, Zhenhua and Qi, Lei and Zhuang, Jiedong and Yang, Wankou},
  journal={IEEE Transactions on Image Processing},
  title={Vision-Based UAV Self-Positioning in Low-Altitude Urban Environments},
  year={2024},
  volume={33},
  number={},
  pages={493-508},
  doi={10.1109/TIP.2023.3346279}}

@inproceedings{r8,
  title={GPS-denied UAV localization using pre-existing satellite imagery},
  author={Goforth, Hunter and Lucey, Simon},
  booktitle={2019 International conference on robotics and automation (ICRA)},
  pages={2974--2980},
  year={2019},
  organization={IEEE}
}

@article{r9,
  title={TransFG: A cross-view geo-localization of satellite and UAVs imagery pipeline using transformer-based feature aggregation and gradient guidance},
  author={Zhao, Hu and Ren, Keyan and Yue, Tianyi and Zhang, Chun and Yuan, Shuai},
  journal={IEEE Transactions on Geoscience and Remote Sensing},
  volume={62},
  pages={1--12},
  year={2024},
  publisher={IEEE}
}

@inproceedings{r10,
  title={APA-BI: Adaptive Partition Aggregation and Bidirectional Integration for UAV-View Geo-Localization},
  author={Zhang, Xichen and Zhao, Shuying and Zhang, Yunzhou and Ge, Fawei and Zhao, Bin and Zhang, Yizhong},
  booktitle={2025 IEEE International Conference on Robotics and Automation (ICRA)},
  pages={4175--4181},
  year={2025},
  organization={IEEE}
}

@article{r11,
  title={SHAA: Spatial Hybrid Attention Network with Adaptive Cross-Entropy Loss Function for UAV-view Geo-localization},
  author={Chen, Nanhua and Zhang, Dongshuo and Jiang, Kai and Yu, Meng and Zhu, Yeqing and Lou, Tai-shan and Zhao, Liangyu},
  journal={IEEE Transactions on Circuits and Systems for Video Technology},
  year={2025},
  publisher={IEEE}
}

@inproceedings{r12,
  title={Bridging the domain gap for ground-to-aerial image matching},
  author={Regmi, Krishna and Shah, Mubarak},
  booktitle={Proceedings of the IEEE/CVF International Conference on Computer Vision},
  pages={470--479},
  year={2019}
}

@article{r13,
  title={UAV-satellite view synthesis for cross-view geo-localization},
  author={Tian, Xiaoyang and Shao, Jie and Ouyang, Deqiang and Shen, Heng Tao},
  journal={IEEE Transactions on Circuits and Systems for Video Technology},
  volume={32},
  number={7},
  pages={4804--4815},
  year={2021},
  publisher={IEEE}
}

@inproceedings{r14,
  title={Cross-view meets diffusion: Aerial image synthesis with geometry and text guidance},
  author={Arrabi, Ahmad and Zhang, Xiaohan and Sultani, Waqas and Chen, Chen and Wshah, Safwan},
  booktitle={2025 IEEE/CVF Winter Conference on Applications of Computer Vision (WACV)},
  pages={5356--5366},
  year={2025},
  organization={IEEE}
}

@article{r15,
  title={Generative adversarial networks},
  author={Goodfellow, Ian and Pouget-Abadie, Jean and Mirza, Mehdi and Xu, Bing and Warde-Farley, David and Ozair, Sherjil and Courville, Aaron and Bengio, Yoshua},
  journal={Communications of the ACM},
  volume={63},
  number={11},
  pages={139--144},
  year={2020},
  publisher={ACM New York, NY, USA}
}

@article{r16,
  title={Denoising diffusion probabilistic models},
  author={Ho, Jonathan and Jain, Ajay and Abbeel, Pieter},
  journal={Advances in neural information processing systems},
  volume={33},
  pages={6840--6851},
  year={2020}
}

@inproceedings{r17,
  title={Unleashing unlabeled data: A paradigm for cross-view geo-localization},
  author={Li, Guopeng and Qian, Ming and Xia, Gui-Song},
  booktitle={Proceedings of the IEEE/CVF Conference on Computer Vision and Pattern Recognition},
  pages={16719--16729},
  year={2024}
}

@article{r18,
  title={Unsupervised Multi-view UAV Image Geo-localization via Iterative Rendering},
  author={Li, Haoyuan and Xu, Chang and Yang, Wen and Mi, Li and Yu, Huai and Zhang, Haijian and Xia, Gui-Song},
  journal={IEEE Transactions on Geoscience and Remote Sensing},
  year={2025},
  publisher={IEEE}
}

@inproceedings{r19,
  title={Structure-from-motion revisited},
  author={Schonberger, Johannes L and Frahm, Jan-Michael},
  booktitle={Proceedings of the IEEE conference on computer vision and pattern recognition},
  pages={4104--4113},
  year={2016}
}

@inproceedings{r21,
  title={Sat2scene: 3d urban scene generation from satellite images with diffusion},
  author={Li, Zuoyue and Li, Zhenqiang and Cui, Zhaopeng and Pollefeys, Marc and Oswald, Martin R},
  booktitle={Proceedings of the IEEE/CVF Conference on Computer Vision and Pattern Recognition},
  pages={7141--7150},
  year={2024}
}

@article{r22,
  title={Diffusionsat: A generative foundation model for satellite imagery},
  author={Khanna, Samar and Liu, Patrick and Zhou, Linqi and Meng, Chenlin and Rombach, Robin and Burke, Marshall and Lobell, David and Ermon, Stefano},
  journal={arXiv preprint arXiv:2312.03606},
  year={2023}
}

@inproceedings{r23,
  title={Multi-weather cross-view geo-localization using denoising diffusion models},
  author={Feng, Tongtong and Li, Qing and Wang, Xin and Wang, Mingzi and Li, Guangyao and Zhu, Wenwu},
  booktitle={Proceedings of the 2nd Workshop on UAVs in Multimedia: Capturing the World from a New Perspective},
  pages={35--39},
  year={2024}
}

@article{r24,
  title={CDM-Net: A Framework for Cross-View Geo-Localization With Multimodal Data},
  author={Zhou, Xin and Yang, Xuerong and Zhang, Yanchun},
  journal={IEEE Transactions on Geoscience and Remote Sensing},
  year={2025},
  publisher={IEEE}
}

@inproceedings{r25,
  title={Prompt-free diffusion: Taking" text" out of text-to-image diffusion models},
  author={Xu, Xingqian and Guo, Jiayi and Wang, Zhangyang and Huang, Gao and Essa, Irfan and Shi, Humphrey},
  booktitle={Proceedings of the IEEE/CVF conference on computer vision and pattern recognition},
  pages={8682--8692},
  year={2024}
}

@inproceedings{r26,
  title={Towards natural language-guided drones: GeoText-1652 benchmark with spatial relation matching},
  author={Chu, Meng and Zheng, Zhedong and Ji, Wei and Wang, Tingyu and Chua, Tat-Seng},
  booktitle={European Conference on Computer Vision},
  pages={213--231},
  year={2024},
  organization={Springer}
}

@inproceedings{r27,
  title={Lending orientation to neural networks for cross-view geo-localization},
  author={Liu, Liu and Li, Hongdong},
  booktitle={Proceedings of the IEEE/CVF conference on computer vision and pattern recognition},
  pages={5624--5633},
  year={2019}
}

@article{r28,
  title={Each part matters: Local patterns facilitate cross-view geo-localization},
  author={Wang, Tingyu and Zheng, Zhedong and Yan, Chenggang and Zhang, Jiyong and Sun, Yaoqi and Zheng, Bolun and Yang, Yi},
  journal={IEEE Transactions on Circuits and Systems for Video Technology},
  volume={32},
  number={2},
  pages={867--879},
  year={2021},
  publisher={IEEE}
}

@article{r29,
  title={A transformer-based feature segmentation and region alignment method for UAV-view geo-localization},
  author={Dai, Ming and Hu, Jianhong and Zhuang, Jiedong and Zheng, Enhui},
  journal={IEEE Transactions on Circuits and Systems for Video Technology},
  volume={32},
  number={7},
  pages={4376--4389},
  year={2021},
  publisher={IEEE}
}

@article{r30,
  title={UAV’s status is worth considering: A fusion representations matching method for geo-localization},
  author={Zhu, Runzhe and Yang, Mingze and Yin, Ling and Wu, Fei and Yang, Yuncheng},
  journal={Sensors},
  volume={23},
  number={2},
  pages={720},
  year={2023},
  publisher={MDPI}
}

@article{r31,
  title={Direction-guided multiscale feature fusion network for geo-localization},
  author={Lv, Hongxiang and Zhu, Hai and Zhu, Runzhe and Wu, Fei and Wang, Chunyuan and Cai, Meiyu and Zhang, Kaiyu},
  journal={IEEE Transactions on Geoscience and Remote Sensing},
  volume={62},
  pages={1--13},
  year={2024},
  publisher={IEEE}
}

@inproceedings{r32,
  title={Content-aware hierarchical representation selection for cross-view geo-localization},
  author={Lu, Zeng and Pu, Tao and Chen, Tianshui and Lin, Liang},
  booktitle={Proceedings of the asian conference on computer vision},
  pages={4211--4224},
  year={2022}
}

@article{r33,
  title={Sdpl: Shifting-dense partition learning for uav-view geo-localization},
  author={Chen, Quan and Wang, Tingyu and Yang, Zihao and Li, Haoran and Lu, Rongfeng and Sun, Yaoqi and Zheng, Bolun and Yan, Chenggang},
  journal={IEEE Transactions on Circuits and Systems for Video Technology},
  volume={34},
  number={11},
  pages={11810--11824},
  year={2024},
  publisher={IEEE}
}

@inproceedings{r34,
  title={Mfrgn: Multi-scale feature representation generalization network for ground-to-aerial geo-localization},
  author={Wang, Yuntao and Zhang, Jinpu and Wei, Ruonan and Gao, Wenbo and Wang, Yuehuan},
  booktitle={Proceedings of the 32nd ACM International Conference on Multimedia},
  pages={2574--2583},
  year={2024}
}

@article{r35,
  title={TirSA: A three stage approach for UAV-satellite cross-view geo-localization based on self-supervised feature enhancement},
  author={Sun, Jian and Sun, Hao and Lei, Lin and Ji, Kefeng and Kuang, Gangyao},
  journal={IEEE Transactions on Circuits and Systems for Video Technology},
  volume={34},
  number={9},
  pages={7882--7895},
  year={2024},
  publisher={IEEE}
}

@article{r36,
  title={MCCG: A ConvNeXt-based multiple-classifier method for cross-view geo-localization},
  author={Shen, Tianrui and Wei, Yingmei and Kang, Lai and Wan, Shanshan and Yang, Yee-Hong},
  journal={IEEE Transactions on Circuits and Systems for Video Technology},
  volume={34},
  number={3},
  pages={1456--1468},
  year={2023},
  publisher={IEEE}
}

@article{r37,
  title={Multi-level embedding and alignment network with consistency and invariance learning for cross-view geo-localization},
  author={Chen, Zhongwei and Yang, Zhao-Xu and Rong, Hai-Jun},
  journal={IEEE Transactions on Geoscience and Remote Sensing},
  year={2025},
  publisher={IEEE}
}

@article{r38,
  title={Query-Driven Feature Learning for Cross-View Geo-Localization},
  author={Hu, Shuyu and Shi, Zelin and Jin, Tong and Liu, Yunpeng},
  journal={IEEE Transactions on Geoscience and Remote Sensing},
  year={2025},
  publisher={IEEE}
}

@article{r39,
  title={Joint Representation Learning Based on Feature Center Region Diffusion and Edge Radiation for Cross-View Geo-Localization},
  author={Ge, Fawei and Zhang, Yunzhou and Wang, Li and Liu, Yixiu and Si, Pengju and Zhang, Jinjin and Shen, You},
  journal={IEEE Transactions on Geoscience and Remote Sensing},
  year={2024},
  publisher={IEEE}
}

@inproceedings{r47,
  title={Deep residual learning for image recognition},
  author={He, Kaiming and Zhang, Xiangyu and Ren, Shaoqing and Sun, Jian},
  booktitle={Proceedings of the IEEE conference on computer vision and pattern recognition},
  pages={770--778},
  year={2016}
}

@inproceedings{r48,
  title={High-resolution image synthesis with latent diffusion models},
  author={Rombach, Robin and Blattmann, Andreas and Lorenz, Dominik and Esser, Patrick and Ommer, Bj{\"o}rn},
  booktitle={Proceedings of the IEEE/CVF conference on computer vision and pattern recognition},
  pages={10684--10695},
  year={2022}
}

@ARTICLE{r50,
  author={Canny, John},
  journal={IEEE Transactions on Pattern Analysis and Machine Intelligence}, 
  title={A Computational Approach to Edge Detection}, 
  year={1986},
  volume={PAMI-8},
  number={6},
  pages={679-698},
  doi={10.1109/TPAMI.1986.4767851}}

@article{r51,
  title={Sam 2: Segment anything in images and videos},
  author={Ravi, Nikhila and Gabeur, Valentin and Hu, Yuan-Ting and Hu, Ronghang and Ryali, Chaitanya and Ma, Tengyu and Khedr, Haitham and R{\"a}dle, Roman and Rolland, Chloe and Gustafson, Laura and others},
  journal={arXiv preprint arXiv:2408.00714},
  year={2024}
}

@article{r52,
  title={Depth anything v2},
  author={Yang, Lihe and Kang, Bingyi and Huang, Zilong and Zhao, Zhen and Xu, Xiaogang and Feng, Jiashi and Zhao, Hengshuang},
  journal={Advances in Neural Information Processing Systems},
  volume={37},
  pages={21875--21911},
  year={2024}
}

@inproceedings{r54,
  title={Cbam: Convolutional block attention module},
  author={Woo, Sanghyun and Park, Jongchan and Lee, Joon-Young and Kweon, In So},
  booktitle={Proceedings of the European conference on computer vision (ECCV)},
  pages={3--19},
  year={2018}
}

@article{r55,
  title={Very deep convolutional networks for large-scale image recognition},
  author={Simonyan, Karen and Zisserman, Andrew},
  journal={arXiv preprint arXiv:1409.1556},
  year={2014}
}

@article{r58,
  title={Lora: Low-rank adaptation of large language models.},
  author={Hu, Edward J and Shen, Yelong and Wallis, Phillip and Allen-Zhu, Zeyuan and Li, Yuanzhi and Wang, Shean and Wang, Lu and Chen, Weizhu and others},
  journal={ICLR},
  volume={1},
  number={2},
  pages={3},
  year={2022}
}

@article{r59,
  title={A practical cross-view image matching method between UAV and satellite for UAV-based geo-localization},
  author={Ding, Lirong and Zhou, Ji and Meng, Lingxuan and Long, Zhiyong},
  journal={Remote Sensing},
  volume={13},
  number={1},
  pages={47},
  year={2020},
  publisher={MDPI}
}

@article{r60,
  title={Multibranch joint representation learning based on information fusion strategy for cross-view geo-localization},
  author={Ge, Fawei and Zhang, Yunzhou and Liu, Yixiu and Wang, Guiyuan and Coleman, Sonya and Kerr, Dermot and Wang, Li},
  journal={IEEE Transactions on Geoscience and Remote Sensing},
  volume={62},
  pages={1--16},
  year={2024},
  publisher={IEEE}
}

@article{r61,
  title={Segcn: A semantic-aware graph convolutional network for uav geo-localization},
  author={Liu, Xiangzeng and Wang, Ziyao and Wu, Yue and Miao, Qiguang},
  journal={IEEE Journal of Selected Topics in Applied Earth Observations and Remote Sensing},
  volume={17},
  pages={6055--6066},
  year={2024},
  publisher={IEEE}
}

@article{r62,
  title={Multilevel feedback joint representation learning network based on adaptive area elimination for cross-view geo-localization},
  author={Ge, Fawei and Zhang, Yunzhou and Wang, Li and Liu, Wei and Liu, Yixiu and Coleman, Sonya and Kerr, Dermot},
  journal={IEEE transactions on geoscience and remote sensing},
  volume={62},
  pages={1--15},
  year={2024},
  publisher={IEEE}
}

@article{r63,
  title={Ccr: A counterfactual causal reasoning-based method for cross-view geo-localization},
  author={Du, Haolin and He, Jingfei and Zhao, Yuanqing},
  journal={IEEE Transactions on Circuits and Systems for Video Technology},
  year={2024},
  publisher={IEEE}
}

@inproceedings{r64,
  title={Sample4geo: Hard negative sampling for cross-view geo-localisation},
  author={Deuser, Fabian and Habel, Konrad and Oswald, Norbert},
  booktitle={Proceedings of the IEEE/CVF International Conference on Computer Vision},
  pages={16847--16856},
  year={2023}
}

@article{r65,
  title={Learning Robust Feature Representation for Cross-View Image Geo-localization},
  author={Gan, Wenjian and Zhou, Yang and Hu, Xiaofei and Zhao, Luying and Huang, Gaoshuang and Hou, Mingbo},
  journal={IEEE Geoscience and Remote Sensing Letters},
  year={2025},
  publisher={IEEE}
}

@article{r66,
  title={Enhancing cross-view geo-localization with domain alignment and scene consistency},
  author={Xia, Panwang and Wan, Yi and Zheng, Zhi and Zhang, Yongjun and Deng, Jiwei},
  journal={IEEE Transactions on Circuits and Systems for Video Technology},
  year={2024},
  publisher={IEEE}
}

@article{r67,
  title={CGSI: Context-Guided and UAV’s Status Informed Multimodal Framework for Generalizable Cross-View Geo-Localization},
  author={Sun, Jian and Huang, Junlang and Jiang, Xinyu and Zhou, Yimin and VONG, Chi-Man},
  journal={IEEE Transactions on Circuits and Systems for Video Technology},
  year={2025},
  publisher={IEEE}
}

@article{r70,
  title={Dual-path convolutional image-text embeddings with instance loss},
  author={Zheng, Zhedong and Zheng, Liang and Garrett, Michael and Yang, Yi and Xu, Mingliang and Shen, Yi-Dong},
  journal={ACM Transactions on Multimedia Computing, Communications, and Applications (TOMM)},
  volume={16},
  number={2},
  pages={1--23},
  year={2020},
  publisher={ACM New York, NY, USA}
}

@inproceedings{r71,
  title={Resolution-robust large mask inpainting with fourier convolutions},
  author={Suvorov, Roman and Logacheva, Elizaveta and Mashikhin, Anton and Remizova, Anastasia and Ashukha, Arsenii and Silvestrov, Aleksei and Kong, Naejin and Goka, Harshith and Park, Kiwoong and Lempitsky, Victor},
  booktitle={Proceedings of the IEEE/CVF winter conference on applications of computer vision},
  pages={2149--2159},
  year={2022}
}

% \newpage
% \begin{IEEEbiography}[{\includegraphics[width=1in,height=1.25in,clip,keepaspectratio]{liutao.png}}]{Tao Liu}
%  is currently pursuing a Ph.D. at Nanjing University of Science and Technology. His research interests include image processing, drone positioning and navigation, computer vision, and more.
% \end{IEEEbiography}

% \begin{IEEEbiography}[{\includegraphics[width=1in,height=1.25in,clip,keepaspectratio]{renkan.png}}]{Kan Ren}
% received his Ph.D. from the University of Surrey in 2011 and is currently a professor at the School of Electronic Engineering and Optoelectronic Technology at Nanjing University of Science and Technology. His research interests include image processing, drone positioning and navigation, and more.
% \end{IEEEbiography}

% \begin{IEEEbiography}[{\includegraphics[width=1in,height=1.25in,clip,keepaspectratio]{chenqian.png}}]{Qian Chen}
% received his Ph.D. from Nanjing University of Science and Technology in 1996 and is currently a professor at the School of Electronic Engineering and Optoelectronic Technology at Nanjing University of Science and Technology. His research interests include image processing, drone positioning and navigation, and more.
% \end{IEEEbiography}

% \vfill

% Biography text here without a photo.

% \begin{IEEEbiography}[{\includegraphics[width=1in,height=1.25in,clip,keepaspectratio]{fig1.png}}]{IEEE Publications Technology Team}
% In this paragraph you can place your educational, professional background and research and other interests.\end{IEEEbiography}

\end{document}